\documentclass[sn-mathphys-num]{sn-jnl}

\usepackage{graphicx}%
\usepackage{multirow}%
\usepackage{dirtytalk}%
\usepackage{amsmath,amssymb,amsfonts}%
\usepackage{amsthm}%
\usepackage{mathrsfs}%
\usepackage{mdframed}
\usepackage[title]{appendix}%
\usepackage{xcolor}%
\usepackage{textcomp}%
\usepackage{manyfoot}%
\usepackage{booktabs}%
\usepackage{algorithm}%
\usepackage{algorithmicx}%
\usepackage{algpseudocode}%
\usepackage{listings}%
\usepackage{makecell}
\usepackage{multirow}
\usepackage{amsmath} 
\usepackage{amsfonts} 
\usepackage{marvosym} 
\usepackage{tabularx}
\usepackage{fontawesome}
\newcolumntype{"}{@{\hskip\tabcolsep\vrule width 2pt\hskip\tabcolsep}}

\setcitestyle{authoryear,open={(},close={)}}
\begin{document}

\title[Article Title]{\textbf{\texttt{FakeWatch} \faEye}: A Framework for Detecting Fake News to Ensure Credible Elections}

\author[1]{\fnm{Shaina} \sur{Raza}}\email{shaina.raza@vectorinstitute.ai}

\author[1]{\fnm{Tahniat} \sur{Khan}}\email{tahniat.khan@vectorinstitute.ai}

\author[1]{\fnm{Veronica} \sur{Chatrath}}\email{veronica.chatrath@vectorinstitute.ai}

\author[3]{\fnm{Drai} \sur{Paulen-Patterson }}\email{draip@live.com}

\author[2]{\fnm{Mizanur} \sur{Rahman}}\email{ mizanur.york@gmail.com}
\author[1]{\fnm{Oluwanifemi} \sur{Bamgbose}}\email{oluwanifemi.bamgbose@vectorinstitute.ai}

\affil[1]{\orgname{Vector Institute}, \orgaddress{\city{Toronto}, \state{Ontario}, \country{Canada}}}

\affil[2]{\orgname{RBC Royal Bank}, \city{Toronto}, \state{Ontario}, \country{Canada}}

\affil[3]{\orgname{Toronto Metropolitan University}, \city{Toronto}, \state{Ontario}, \country{Canada}}


\abstract{In today's technologically driven world, the rapid spread of fake news, particularly during critical events like elections, poses a growing threat to the integrity of information. To tackle this challenge head-on, we introduce \textbf{\texttt{FakeWatch} \faEye}, a comprehensive framework carefully designed to detect fake news. Leveraging a newly curated dataset of North American election-related news articles, we construct robust classification models. Our framework integrates a model hub comprising of both traditional machine learning (ML) techniques, and state-of-the-art Language Models (LMs) to discern fake news effectively. Our objective is to provide the research community with adaptable and precise classification models adept at identifying fake news for the elections agenda. Quantitative evaluations of fake news classifiers on our dataset reveal that, while state-of-the-art LMs exhibit a slight edge over traditional ML models, classical models remain competitive due to their balance of accuracy and computational efficiency. Additionally, qualitative analyses shed light on patterns within fake news articles. We provide our labeled data\footnote{\url{https://huggingface.co/datasets/newsmediabias/fake_news_elections_labelled_data} }~ and model\footnote{\url{https://huggingface.co/newsmediabias/FakeWatch} }~ for reproducibility and further research.}




\maketitle

\section{Introduction}\label{sec1}

Fake news encompasses false or misleading information presented as if it were true, with the intent to deceive or manipulate. This deceptive practice is achieved through misinformation, which is spread without intent, or through disinformation, which is deliberately created to mislead the reader \citep{zhou_survey_2020,raza_fake_2022}. Fake news spreads through channels including traditional media, social media, websites, and online platforms, encompassing fabricated stories, distorted facts, sensationalized headlines, and selectively edited content. The motivation for creating and spreading misinformation varies from financial gain, to advancing a specific agenda, to acting as a tool for propaganda and influencing public opinion, to sowing confusion.

The negative consequences of fake news are evident in real-world scenarios. During the 2022 Ukraine-Russia conflict, termed \say{World War Wired} for its heavy social media documentation \citep{wright_portrayals_2023}. During the 2020 Munich Security Conference, the Director-General of the World Health Organization (WHO) remarked, \say{We are not just fighting an epidemic; we are fighting an infodemic}, underscoring the alarming speed at which COVID-19 misinformation was spreading, compared to the virus itself \citep{muhammed_t_disaster_2022}. Fake news is also highlighted by a NewsGuard report, showing that new TikTok users quickly encounter misleading war content \citep{russiaUkraine}. 
 Over 6,000 people were hospitalized in the first three months of 2020 due to COVID-19 misinformation, with false claims of microchip implants contributing to vaccine hesitancy \citep{VaccineMyths}. 

Fake news is not just limited to the aforementioned events; it also impacts various aspects of society. Whether during elections, public health crises, or conflicts between nations, combating fake news has become crucial, emphasizing the role of artificial intelligence (AI) classifiers in identifying and mitigating misinformation. Integrating advanced technology and collaborative efforts is essential for navigating the information landscape and fostering a resilient, informed society.

Our research in fake news detection advances existing transformer-based model studies, particularly those by \cite{raza_fake_2022} and \cite{alghamdi_towards_2023}, with a focus on the dynamic context of North American elections. Previous insights from \cite{kaliyar_fakebert_2021, aimeur_fake_2023}, and \cite{ hamed_review_2023} highlight deep learning (DL) techniques, but reveal shortcomings in handling data and concept drift \citep{raza2019news}. Data drift, the performance decline of AI models with new data, and concept drift, the evolution of data patterns, pose challenges in maintaining classifier accuracy, especially critical in the rapidly changing arena of election news. 
In this work we assess the effectiveness of our methodology in detecting fake news during North American elections, considering the need for a new dataset for this purpose.

We present three significant contributions that form the core of our work:

\begin{enumerate}

    \item  We construct a novel 2024 US Elections dataset, curated using targeted keywords and themes, and annotated through a combination of large language models (LLMs) and verified through human reviews. This dataset addresses the absence of available data for building effective fake news classifiers for the 2024 Presidential US Elections. Data prepared for this paper is made available. 
    
    \item  We present an extensive range of machine learning (ML) and DL classifiers for fake news, creating a versatile and comprehensive hub. We release our classification model, trained on our data, publicly for research.

    \item We showcase the construction of data curation and models and evaluated the models through both a qualitative and quantitative lens, for fake news detection. We show specific linguistic patterns—emotional tone, pronoun usage, cognitive complexity, and temporal orientation to distinguish between fake and real news. 
\end{enumerate}

The novelty of our approach lies is its all-encompassing framework that integrates data collection, annotation, and model construction smoothly, for providing a strong basis for detecting fake news regarding, specifically for elections.

\section{Related Works}
Fake news detection is a subtask of text classification \citep{liu_two-stage_2019}, commonly defined as the task of classifying news as either real or false. The term \say{fake news} refers to false or misleading information presented as authentic, with the intention to deceive or mislead people \citep{lu_fake_2022,zhou_survey_2020,allcott2017social,raza_fake_2022}. Fake news takes various forms, including clickbait (misleading headlines), disinformation (intentionally misleading the public), misinformation (false information regardless of motive), hoax, parody, satire, rumor, deceptive news, and others, as discussed in \cite{zhou_survey_2020}.

The recent developments in fake news classification highlight the integration of ML and DL techniques to improve accuracy and efficiency. A recent DL-focused study  \cite{lu_fake_2022} shows how these methods can enhance classifier performance for the fake news detection task. Another related work \citet{arora_reviewing_2023} provides a critical analysis of various algorithms used for fake news classification, shedding light on their effectiveness. On the similar note, a work on the fake news detection \citet{bonny_detecting_2022} demonstrates the potential of ML classifiers in identifying fake news within English news datasets. Some works focus on detecting political news and fake information \citep{raza_automatic_2021} through DL. All these studies represent state-of-the-art efforts to combat misinformation with AI. However, a key challenge in building an AI-based fake news classifier is related to datasets, feature representation, and data fusion \citep{hamed_review_2023}.

There are also many other challenges related to selecting relevant features, considering the dynamic nature of fake news content, such as the temporal evolution of linguistic patterns \citep{raza2019news}, the influence of multimedia \citep{qi2019exploiting}, and the contextual nuances that contribute to more accurate feature representation \citep{fakeNewsData}. Some works \cite{raza_fake_2022,fakeNewsData} recognize the difficulty in identifying fake news due to the interplay between content and social factors, and propose a transformer-based approach, focusing on news content and social contexts.

\textbf{Datasets}: The quality and diversity of datasets used for training and evaluating fake news detection models is a key challenge. Ensuring datasets are representative is crucial for models to generalize to real-world scenarios \citep{raza2023constructing}.  A comprehensive and balanced dataset is required to capture the multifaceted nature of fake news \citep{pubmed}. Benchmark datasets like FakeNewsNet \citep{shu2020fakenewsnet}, Fakeddit \citep{nakamura2019r}, and NELA-GT \citep{gruppi_nela-gt-2022_2023} are mostly used for fake news detection.
In response to the issue of COVID-19-related misinformation, \citet{alghamdi_towards_2023} implemented transformer-based models for fake news detection. Elections and fake news have always been a topic of interest, and many datasets have been proposed for this purpose \citep{allcott2017social,grinberg2019fake,gruppi_nela-gt-2022_2023}. Most of the existing fake news datasets are from the 2016 and 2022 US elections and may not be able to combat data and concept drifts in the 2024 US elections, which led us to develop a new dataset along this line.

Natural Language Processing (NLP) methodologies are fundamental in fake news detection, utilizing datasets such as NELA-GT \cite{gruppi_nela-gt-2022_2023}, Credbank \cite{mitra2015credbank}, FakeNewsNet \cite{shu2020fakenewsnet}, and Liar \cite{wang2017liar} for classification.  Advancements in fake news detection include the analysis of linguistic features using NLP tools, as seen with the WELfake \cite{verma2021welfake} dataset that incorporates pre-processed linguistic features to improve classification accuracy through sentence structure, readability, sentiment analysis, and other linguistic patterns \cite{qi2019exploiting,gaillard2021countering}. 

Exploring image-based and multimodal datasets underscores the complexity of fake news detection, acknowledging that images play a significant role alongside text. Datasets like Image Manipulation and PS-Battles \cite{heller2018ps} are tailored for training models on image-based classification. The exploration of multimodal datasets further highlights the complexity of fake news, emphasizing the need for diverse detection mechanisms \cite{bang2023multitask}. 

Data fusion is crucial to enhancing fake news detection systems. By integrating diverse modalities like textual content \citep{fakeNewsData}, user metadata \citep{raza_fake_2022}, network structures \citep{aimeur_fake_2023}, and language models (LMs) \citep{kaliyar_fakebert_2021}, we can enhance fake news detection robustness. Multimodal detection methods \citep{yang2023multimodal}, are also pivotal as they analyze correlations between image regions and text. Our approach, influenced by these advancements, focuses on optimized data and feature representations for more effective detection.

Table \ref{tab:summary_findings} presents a comparison of our approach with some of the aforementioned recent works.

\begin{table}[h]
    \centering
        \caption{A concise summary of findings of the works mentioned in the Related Works section.}
    \begin{tabular}{p{5cm}|p{6cm}}
        \textbf{Paper}& \textbf{Main Contribution}\\
        \Xhline{2\arrayrulewidth}  \Xhline{2\arrayrulewidth}

        \href{https://www.sciencedirect.com/science/article/abs/pii/S0957417420303274}{Fake news detection in multiple platforms and languages} & Fake news detection across three languages and two platforms using platform/language. \citep{multiplePlatforms} \\
        \hline
        \href{https://link.springer.com/chapter/10.1007/978-3-030-42699-6_9}{Credibility-Based Fake News Detection}  & Credbility of news articles using ML methods. \citep{credibilityBased} \\
        \hline
        \href{https://www.mdpi.com/2079-9292/10/11/1348}{Sentiment Analysis for Fake News Detection} &  Sentiment analysis on fake news using ML methods.  \citep{SentimentAnalysis} \\
        \hline
        \href{https://link.springer.com/article/10.1007/s41060-021-00302-z}{Fake news detection based on news content and social contexts: a transformer-based approach} & Fake news detection using dual transformer-based models. \citep{raza_fake_2022} \\
        \hline
        \href{https://link.springer.com/article/10.1007/s11042-022-12668-8}{Evaluating the effectiveness of publishers’ features in fake news detection on social media} & CreditRank algorithm and a framework for multi-modal fake news detection.  \citep{evaluatingEffect} \\
        \hline
        \href{https://link.springer.com/article/10.1007/s40747-023-01098-0}{Fake news detection based on a hybrid BERT and LightGBM models} & BERT with LightGBM for improved fake news detection accuracy. \citep{HybridBert} \\
        \hline
        
        Our Work (2024) & Framework, benchmarking, and a novel dataset for 2024 Presidential election credibility. \\
        \bottomrule
    \end{tabular}
    \label{tab:summary_findings}
\end{table}

\section{Methodology}

\subsection{Defining Fake News}
Fake news detection can be defined as a binary classification problem, where each news item is assigned a label indicating its correctness. Mathematically, this can be represented as follows:

Let $N$ be a set of news items, where each news item, $n \in N$, is represented by a feature vector, $x_n$, containing its characteristics (e.g., text, metadata, etc.).
There is a corresponding label, \(y_{n}\), for each news item, \emph{n}, where \(y_{n} \in \{ 0,1\}\).

\(y_{n} = 1\) means the news item is fake (false information).

\(y_{n} = 0\) means the news item is real (true information).

A classification model, \emph{f}, is trained to map the feature vector, \(x_{n}\), to its corresponding label, \(y_{n}\) (i.e., vector
\({f(x}_{n}) \rightarrow y_{n}\)). The goal of this model is to
accurately predict the label for each news item based on its features,
thus distinguishing between fake and real news.

\subsection{\texttt{FakeWatch} \faEye~Framework}
\label{fakewatch-electionshield}
\begin{figure*}[h!]
    \centering
    \includegraphics[width=\linewidth]{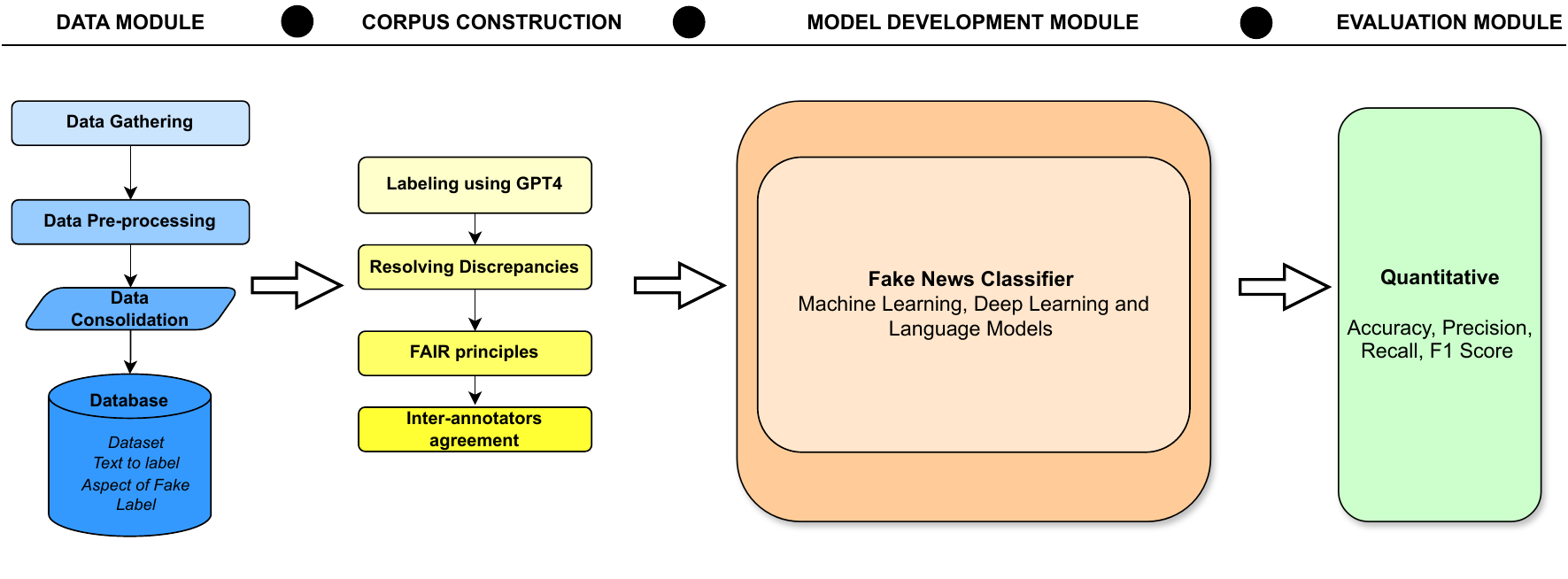}
    \caption{\textbf{\texttt{FakeWatch} \faEye}, a framework to detect biases within textual data. It is a four-module framework, where data is first gathered from diverse sources and then constructed into a quality-focused corpus. Various ML models are trained on the data and evaluated based on different evaluation metrics.}
    \label{fig:fig1}
\end{figure*}

We present \textbf{\texttt{FakeWatch} \faEye}, a framework designed to detect fake news, as illustrated in Figure \ref{fig:fig1}. This framework is structured into four distinct modules: (i) the data collection module, (ii) the corpus construction module, (iii) the model development module, and (iv) the evaluation module. Each module is designed to integrate with the others, providing an effective and comprehensive approach for detecting fake news in texts.

\subsubsection{Data Module}

In this module  we integrate data from two distinct sources: 1) Google RSS for data curation and 2) NELA-GT-2022 dataset \citep{gruppi_nela-gt-2022_2023}, an existing benchmark.

\textit{Data Curation:} We curated data from Google RSS by carefully selecting keywords, categorized into groups such as race/ethnicity-related terms, religious terms, geographical references, historical and political events, and other terms associated with racial discourse. From the Google RSS feeds, we curate the data using the Newspaper3k Python package\footnote{https://newspaper.readthedocs.io/en/latest/} to gather and categorize a wide array of news data from the US over a six-month period (Apr. 20, 2023 - Oct. 20, 2023). This gave us around 50k data based on search query, however, for this work, we labeled around 9000 data points, selecting a significant sample for labeling covering main electoral topics (preferring quality of labels over quantity).
Additionally, we also use the NELA-GT-2022 dataset, which is a benchmark dataset that provides source-level labels for each news article. From the NELA-GT-2022 data 5000 records were filtered in chronological order from Oct. 2022 - Dec. 2022.
Our search query is as follows:
\vspace{0.5cm}

\begin{mdframed}[backgroundcolor=olive!5, linewidth=1pt, linecolor=black, roundcorner=10pt, leftmargin=10, rightmargin=10, innerleftmargin=15, innerrightmargin=15, innertopmargin=15, innerbottommargin=15]
\textbf{Search Query:}
(\say{2024 U.S. elections} OR \say{presidential candidates 2024}) AND \\
(\say{race relations} OR \say{ethnic diversity} OR \say{racial discourse}) AND \\
(\say{religious freedom} OR \say{religious discrimination}) AND \\
(\say{political events} OR \say{historical events} OR \say{geographical impact}) AND \\
(\say{voting patterns} OR \say{electoral college} OR \say{campaign strategies}) AND \\
(\say{news analysis} OR \say{media coverage}) \\
DATES: [2023-04-20 to 2023-10-20] \\
SOURCE: News Feeds
\end{mdframed}

\vspace{0.5cm}


We consolidate data from both sources (curated and NELA-GT) that contains the following columns:
\begin{itemize}
    \item \textit{Dataset:} Specifies the source dataset (e.g., news sources like BBC, CNN, etc.),
    \item \textit{Text: }Contains the actual textual data extracted from the respective datasets,
    \item \textit{Label:} Indicates whether the text is Fake (1) or Real (0), serving as the target variable for the token classifier and for evaluation purposes.
\end{itemize}

Further pre-processing is conducted to prepare the data for subsequent modules of the framework, particularly NLP model, performing token classification. We improved data for ML algorithms by tokenizing and handling missing data. Tokenization breaks text into meaningful units, aiding semantic understanding. Managing missing data prevents bias and boosts model performance. These preprocessing steps create structured data for the NLP token classification model. To safeguard users' privacy, we do not collect user IDs, and employ tokenization to replace references to usernames, URLs, and emails, ensuring all personally identifiable information is private. 

The consolidated dataset underwent a data labelling process, discussed below.

\subsubsection{Corpus Construction}
\label{corpus-construction}

\textit{Data Labelling:} The NELA-GT-2022 dataset includes source-level labels (e.g., BBC, CNN, The Onion), reflecting potential biases associated with news sources. Therefore, it is essential to annotate each individual news article. Similarly, the Google RSS curated data lacks labels for fake news detection, necessitating a strategy for news article annotation.
So, to get the labels, we employed OpenAI's GPT-4 \cite{gpt4} for initial labelling, assessing the likelihood of each news item being fake or real based on its language understanding capabilities. The use of data annotation through GPT-based models is also supported in recent literature \citep{gilardi2023chatgpt}, where the studies show that LLM-based labeling shows better results in terms of less biases and accuracy than crowdsourcing. 

Our annotation process combines automated preliminary assessments by GPT-4 with subsequent manual reviews by trained annotators. This dual approach leverages the efficiency of AI while incorporating human oversight to correct any anomalies and verify the nuanced interpretation of content. Therefore, the likelihood of misclassification is significantly reduced, enhancing the overall trustworthiness of the dataset.

To ensure the reproducibility and credibility of our classification models, we outline comprehensive labeling guidelines and protocols. Each news article undergoes a detailed categorization process as either fake or non-fake, based on justified criteria. These guidelines were developed through consultations with media experts and linguistic analysts to establish a robust framework for fake news detection. 
Our criteria for categorizing articles as fake or non-fake include:

\begin{itemize}
    \item \textcolor{blue}{Source Credibility}: Articles from sources with a history of factual inaccuracies are scrutinized more intensely.
    \item \textcolor{blue}{Linguistic Quality:} Articles with numerous spelling and grammatical errors, or those using sensational language, are flagged for further review.
    \item \textcolor{blue}{Fact-checking}: Statements of fact within articles are cross-verified with trusted databases and news sources.
    \item \textcolor{blue}{Contextual Consistency}: The article's content is checked for consistency with known facts and chronological data.
    \item \textcolor{blue}{Editorial Bias}: Articles are analyzed for potential bias in how information is presented, including the omission of key facts or one-sided reporting.
\end{itemize}

In the consolidated dataframe, each row represents a unique sample from the original dataset, supplying information for fake news detection and assessment. The final combined curated and NELA-GT-2022 dataset comprises a total of 9513 entries, with two unique labels for classification: REAL news, accounting for 5790 entries, and FAKE news, with 4723 entries.


\textit{Data Quality:} 
Recognizing the critical importance of data integrity, we adopted a rigorous approach to ensure the highest quality of our dataset. We engaged a diverse team of six experts—comprising ML Scientists, Data Scientists, Linguistic Experts, and advanced students—for the manual verification of all 9513 records. 
To maintain the highest standards of reliability and consistency in our data labeling process, we implemented strict protocols. Each record underwent an independent review by two experts. Their agreement was quantitatively measured using Cohen’s Kappa coefficient, achieving a score of 0.79. This score, indicative of \say{almost perfect} agreement, confirms the exceptional uniformity and precision of our annotations, thereby reinforcing the credibility and trustworthiness of our dataset.

\subsubsection{Model Development Module}

In this model development module, we establish a comprehensive hub for fake news classification, encompassing traditional ML algorithms and advanced DL models, which are also transformer-based models, including the LMs. Our objective is to showcase the strengths of these diverse approaches, improving the accuracy and efficiency of fake news detection. We aim to deliver a robust and adaptable solution to combat misinformation, offering valuable insights into the real-world performance and scalability of these methods.

\begin{figure}[h!]
    \centering
    \includegraphics[width=1\linewidth]{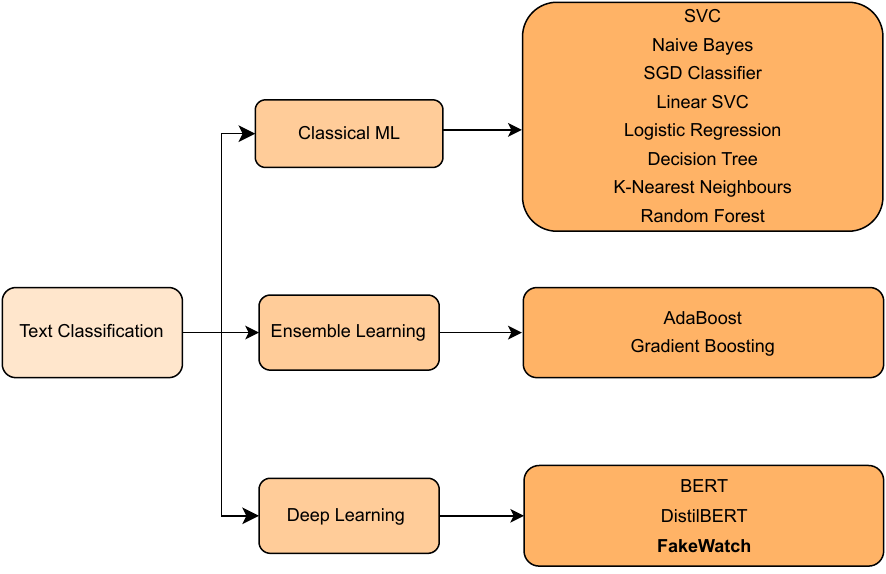}
    \caption{The chosen classification methods.}
    \label{fig:classificationModels}
\end{figure}

To facilitate a structured comparison, we categorize different models into three distinct groups, as depicted in Figure \ref{fig:classificationModels}. 
The models considered for comparison are strategically grouped ML models, advanced DL models, such as transformer-based models and our carefully fine-tuned \textbf{\texttt{FakeWatch} \faEye}. The details of our \textbf{\texttt{FakeWatch} \faEye} is below and brief details for other models are given next:

\noindent \textbf{\texttt{FakeWatch} \faEye }: We have designed \textbf{\texttt{FakeWatch} \faEye }, a specialized LLM derived from the RoBERTa architecture \citep{liu2019roberta},  fine-tuned with our curated dataset. To integrate the RoBERTa model for fake news detection, our methodology is as follows:

Initially, input data undergoes tokenization via Byte-Pair Encoding (BPE), a pivotal preprocessing step for RoBERTa. This involves splitting text into subword units, appending the special tokens, \texttt{[CLS]} at the beginning, and \texttt{[SEP]} at the end, for classification tasks, and segment separation, respectively. Each token is mapped to a high-dimensional space to produce token embeddings. Additionally, position embeddings are added to retain sequential information, crucial for understanding the text structure.

The core of RoBERTa's architecture is the attention mechanism, which allows the model to focus on different parts of the input sequence when predicting an output. The scaled dot-product attention is computed as:
\begin{equation}
\text{Attention}(Q, K, V) = \text{softmax}\left(\frac{QK^T}{\sqrt{d_k}}\right)V
\end{equation}
where $Q$, $K$, and $V$ are the query, key, and value matrices, respectively, and $d_k$ is the dimensionality of the key.

RoBERTa utilizes multiple modules of transformer blocks, where each block comprises two main components: a multi-head self-attention mechanism and a position-wise fully connected feed-forward network. Layer normalization and residual connections are employed around each of these components. The output from the last transformer block passes through a linear layer, followed by a softmax function to predict the probability distribution over the classes. For binary classification (fake vs. real news), the model outputs the probability of the input being in either category.

Finally, the model is fine-tuned on our labeled dataset specific to the task of fake news detection. This involves adjusting the pre-trained weights to better suit the nuances of the fake news classification problem.

\noindent \textbf{Models in the Hub}

\noindent \textbf{Naive Bayes}: A probabilistic classifier based on Bayes' theorem, assuming strong independence between features. Simple and effective, especially in text classification tasks like spam detection.

\noindent \textbf{Logistic Regression:} A fundamental statistical model that
predicts the probability of a binary outcome based on input features. It is widely used for binary classification problems,
such as credit scoring and medical diagnosis.

\noindent \textbf{SGD (Stochastic Gradient Descent) Classifier:} A linear classifier (like Support Vector Machine (SVM) or
logistic regression) that uses gradient descent to optimize the loss function. Ideal for large-scale and sparse ML problems.

\noindent \textbf{Random Forest:} An ensemble learning method that constructs
multiple decision trees at training time and outputs the mode of the
classes (classification) of the individual trees. Effective for handling a large dataset with high dimensionality.

\noindent \textbf{SVC (Support Vector Classifier)}: Part of the SVM family, it is used for classification problems. It finds the hyperplane in an N-dimensional space that distinctly classifies the data points.

\noindent \textbf{Linear SVC:} Similar to SVC, with the parameter kernel set to \textit{linear}, but implemented in terms of liblinear rather than libsvm, so there is more flexibility in the choice
of penalties and loss functions and should scale better to large numbers of samples.

\noindent \textbf{Decision Tree:} A tree-like model of decisions. A type of supervised learning algorithm (having a pre-defined target variable) used in statistics, data mining, and ML.

\noindent \textbf{AdaBoost:} Short for Adaptive Boosting, it is an ensemble technique that combines multiple weak classifiers to create a strong classifier. Effective at improving the accuracy of any given learning algorithm.

\noindent \textbf{Gradient Boosting:} Another ensemble technique that builds trees in a sequential manner, where each tree tries to correct the errors made by the previous one. It is used widely for both regression and classification problems.

\noindent \textbf{DistilBERT:} A smaller, faster, cheaper, and lighter version of BERT. It distills the crucial information from BERT, retaining 97\% of its language understanding capabilities, but with a lower computational cost.

\noindent \textbf{BERT (Bidirectional Encoder Representations from Transformers):} A transformer-based ML technique for NLP pre-training. It is designed to understand the context of a word in a sentence, which significantly improves the state-of-the-art in sentence understanding.

\noindent \textbf{Llama 2-7B:} We employ the Llama-2-7B chat \cite{touvron2023llama} model developed by Meta Platforms, which is a text generation model based on the Transformers architecture using PyTorch. We operate this model in few-shot settings through inference without the need for retraining. We provide prompts to the model, use the generated labels as predictions, and compare these with the ground truth labels from the test set.

Each of these models brings its unique strengths to the task of fake news detection, allowing for a comprehensive approach that enhances the robustness and accuracy of our classification system. 


\subsubsection{Evaluation Module}

The evaluation module is important in assessing the robustness and efficacy of our \textbf{\texttt{FakeWatch} \faEye}~model, employing a multifaceted approach that combines quantitative and qualitative evaluation methods. Quantitatively, we utilize a suite of metrics to comprehensively measure the model's performance. Qualitatively, our evaluations consists of some tests on model's real-world applicability and performance in complex scenarios, including analysis of the linguistic patterns of fake news, topics in the data, and semantic analysis.



\section{Experiments}
\label{experiments}

  \subsection{Settings and Hyperparameters}

We utilize a hardware setup comprising an Intel(R) Core(TM) i7-8565U CPU for local processing. For enhanced computational capabilities, we leverage Google Colab Pro equipped with cloud-based GPUs, enabling efficient execution of resource-intensive tasks. Storage solutions are facilitated through Google Drive, ensuring seamless access to datasets and model checkpoints. On the software, we employ PyTorch BERT, a powerful framework provided by Hugging Face, to implement BERT encoder layers. This allows us to leverage state-of-the-art NLP capabilities for our tasks. Furthermore, to ensure the efficacy of our evaluation strategies, human assessment is incorporated into the process. This involves expert evaluation and validation of the model outputs, providing valuable insights into its performance and effectiveness in real-world scenarios. General hyperparameters for all models in our experiments can be seen in Table \ref{hyperparams}.
We adopt an 80-20 split for training and testing across all models to maintain consistency. To mitigate the issue of data imbalance, we implement an upscaling technique, ensuring equal representation of both classes in our training set. 

\begin{table}[h!]
\centering
\caption{Table of Hyperparameters.}
\label{hyperparams}
\begin{tabular}{|p{2.5cm}|p{8cm}|}
\hline
\textbf{Model} & \textbf{Hyperparameters} \\
\Xhline{2\arrayrulewidth}  \Xhline{2\arrayrulewidth}
Naive Bayes                            & Alpha: 1.0, Fit Prior: \texttt{True} \\
\hline
Logistic Regression                    & C: 1.0, Solver: \texttt{lbfgs}, Penalty: \texttt{l2} \\
\hline
SGD Classifier                         & Loss Function: \texttt{hinge}, Penalty: \texttt{l2}, Learning Rate: 0.01  \\
\hline
Random Forest                          & Number of Trees: 100, Max Depth: \texttt{None}, Max Features: \texttt{auto}  \\
\hline
AdaBoost                               & Number of Estimators: 50, Learning Rate: 1.0     \\
\hline
Gradient Boosting                      & Learning Rate: 0.1, Number of Estimators: 100, Max Depth: 3   \\
\hline
SVC                                    & Kernel: \texttt{rbf}, C: 1.0, Gamma: \texttt{scale}  \\
\hline
Linear SVC                             & C: 1.0, Loss: \texttt{squared\_hinge}, Penalty: \texttt{l2} \\
\hline
Decision Tree                          & Max Depth: None, Min Samples Split: 2, Criterion: \texttt{gini}   \\
\hline
DistilBERT                             & Learning Rate: 6$\times 10^{-6}$, Number of Epochs: 8, Batch Size: 32 \\
\hline
BERT                                   & Learning Rate: 6$\times 10^{-6}$, Number of Epochs: 8, Batch Size: 32  \\
\hline
 Llama 2-7B-chat&Two shot demonstrations (examples) with a temperature of 0.7, maximum token limit of 512, Top-k sampling of 40, and Top-p (nucleus) sampling of 0.9.\\
\hline
\textbf{\texttt{FakeWatch} \faEye }& Learning Rate: 5$\times 10^{-6}$, Number of Epochs: 4, Batch Size: 16, Warmup Steps: 500, Weight Decay: 0.01 \\
\hline
\end{tabular}
\end{table}

\subsection{Evaluation Metrics}

\paragraph{Quantitative Measures}
We use the following evaluation metrics that are commonly used for comparing classification models:


\begin{equation}
    Accuracy = \frac{TP + TN}{TP + TN + FP + FN}
\end{equation}

\begin{equation}
Precision = \frac{TP}{TP + FP}
\end{equation}

\begin{equation}
Recall = \frac{TP}{TP + FN}
\end{equation}

\begin{equation}
F1\ score = 2 \times \frac{Precision \times Recall}{Precision + Recall}
\end{equation}
\vspace{3mm}

\noindent where TP stands for true positive, TN for true negatives, FN for false positives, and FP for false positives.

AUC (Area Under the Curve) provides a scalar measure of the model's ability to discriminate between classes at various threshold settings, with higher values indicating better classification performance. The ROC curve itself plots the true positive rate (TPR) against the false positive rate (FPR) at different threshold levels, providing a graphical representation of the model's classification ability. 

\paragraph{Qualitative Analysis}
We perform a qualitative analysis by incorporating the exploration of linguistic patterns in the data that contribute to fake news. We also perform topic modeling and social network analysis to explore the thematic patterns and the interconnectedness of topics within our corpus. This comprehensive approach not only confirms our hypotheses with statistical evidence but also enriches our understanding of the linguistic patterns and narrative structures characteristic of fake news.
\subsection{Exploratory Analysis on Dataset} 



An exploratory analysis was conducted on the consolidated dataset, and we present here the key insights for conciseness.

\begin{figure}[h]
    \centering
    \includegraphics[width=1\linewidth]{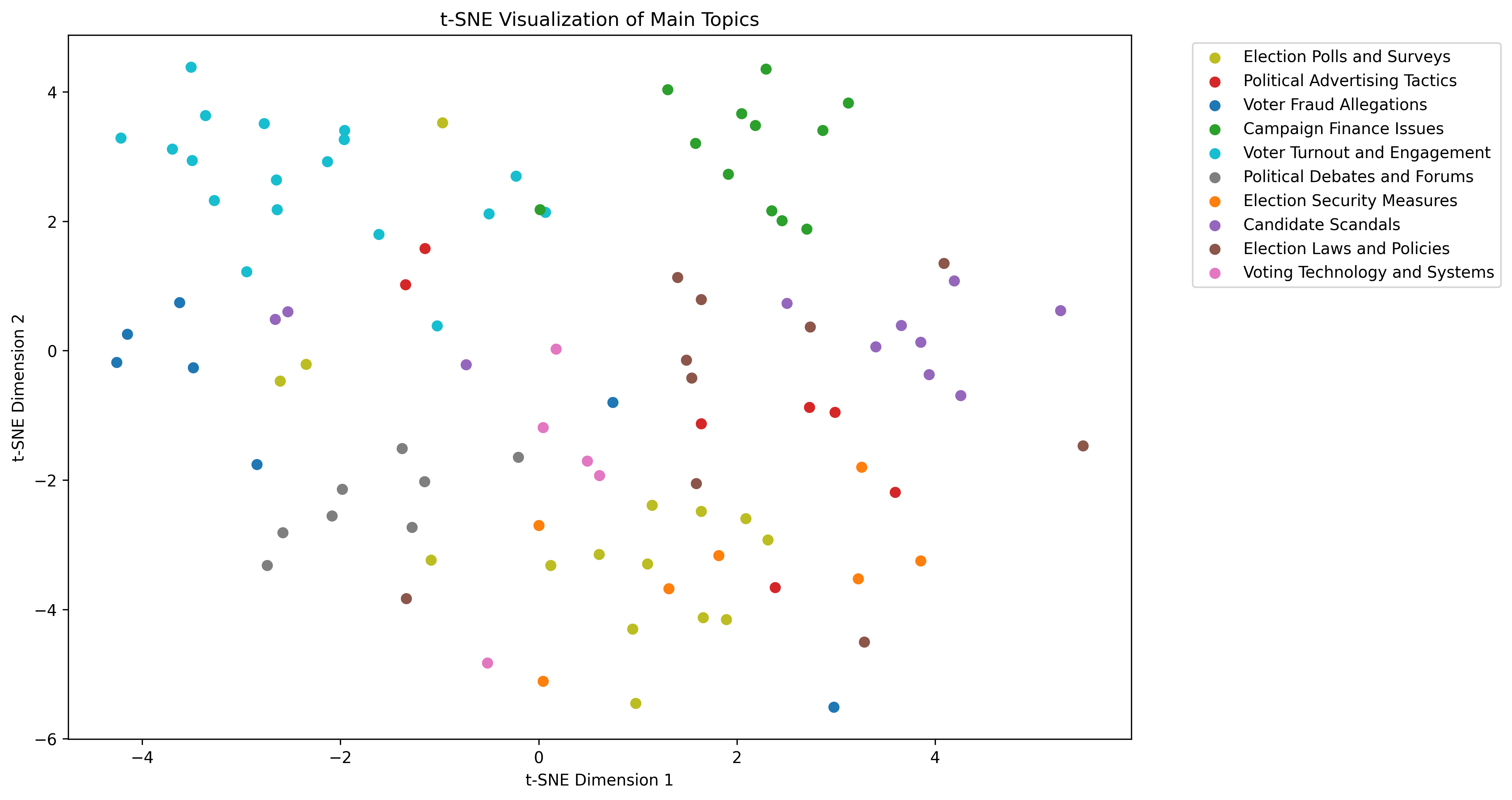}
    \caption{Important topics extracted from the corpus. Each point represents a document, and the color of the point indicates its most dominant topic, labelled according to the legend. Similar content clusters are based on dominant topics, and different topics are positioned farther apart.}
    \label{fig:topicsVis}
\end{figure}

In Figure \ref{fig:topicsVis}, we use t-SNE (t-Distributed Stochastic Neighbor Embedding) on a subset of our data associated with 30 different election-related topics. Figure \ref{fig:topicsVis} shows that topics such as elections, politics, votes, and campaigns are closely placed.

\begin{figure}
    \centering
    \includegraphics[width=1\linewidth]{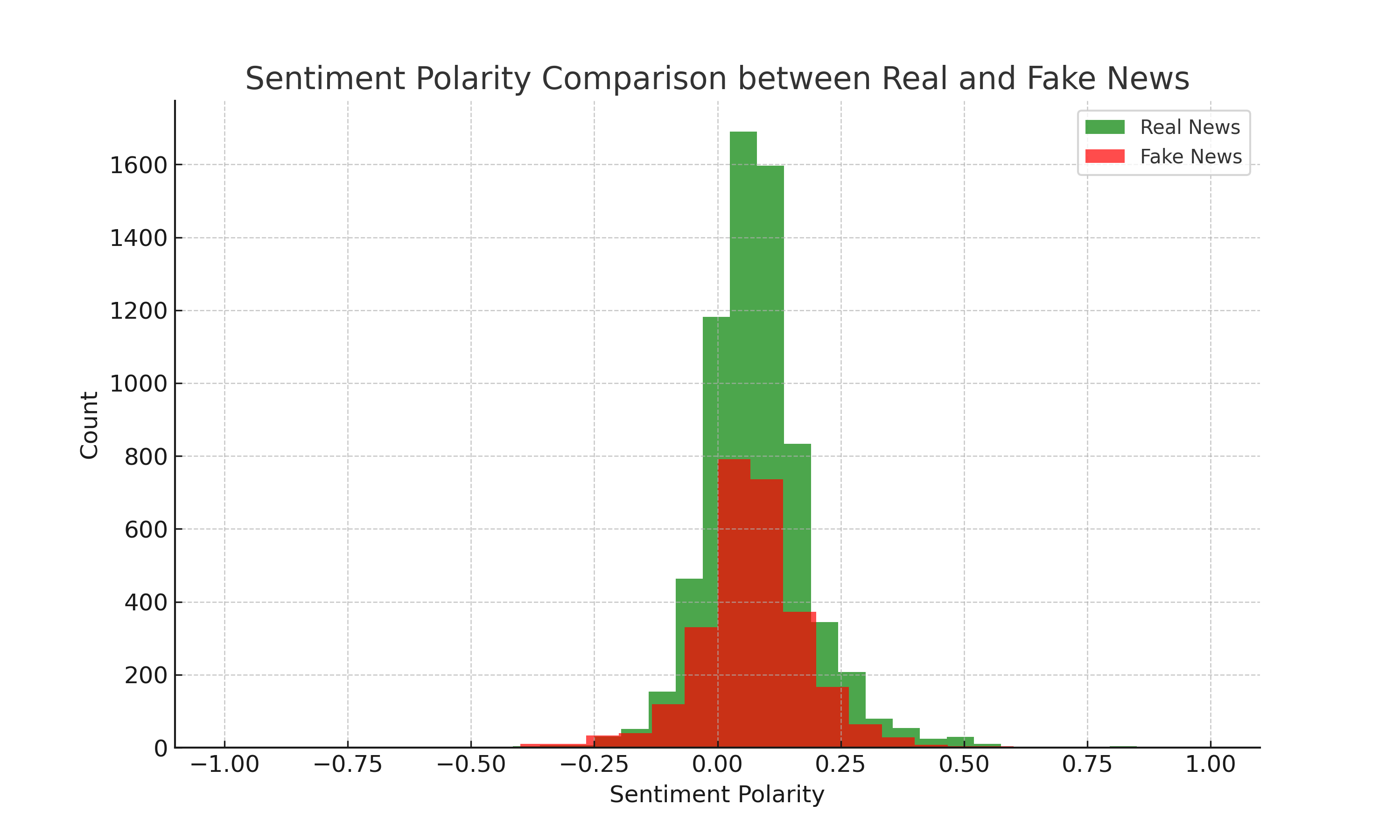}
    \caption{A histogram of sentiment polarity comparison between real (green) and fake (red) news.}
    \label{fig:sentiment}
\end{figure}

We performed sentiment analysis on the news articles to assess the emotional tone conveyed through their text. Using the TextBlob\footnote{https://textblob.readthedocs.io/en/dev/} tool, we calculate the sentiment polarity scores, which range from -1 (extremely negative) to 1 (extremely positive). For each article, we compared the distribution of these scores between real and fake news. The resulting histogram in Figure \ref{fig:sentiment} provides insights into the emotional undertones associated with each type of news, revealing whether fake news articles tend to have a more negative, positive, or similar sentiment compared to real news. The finding from this figure shows that real and fake news articles exhibit similar sentiment distributions, with a slight tendency towards neutral to positive tones, which indicates that sentiment alone may not be sufficient to differentiate between them.

\begin{figure}
    \centering
    \includegraphics[width=1\linewidth]{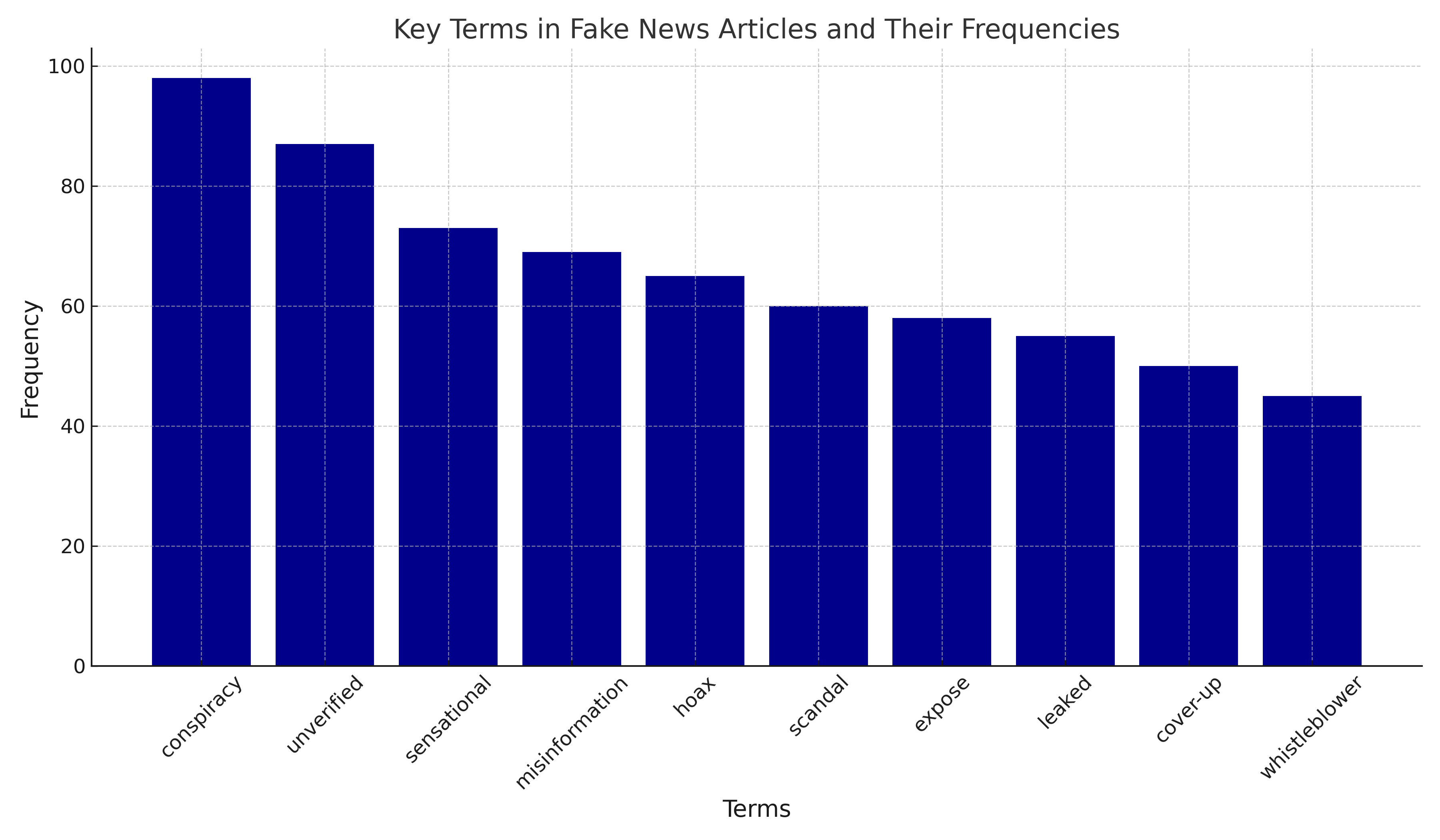}
    \caption{A bar chart of the frequency of each key term in the data.}
    \label{fig:keyTermFigure}
\end{figure}

We utilized the TF-IDF (Term Frequency-Inverse Document Frequency) method to calculate the significance of words within the fake news articles compared to their distribution across the entire dataset. The result, shown in Figure \ref{fig:keyTermFigure}, showcases the frequency of these key terms, with terms like \say{conspiracy}, \say{unverified}, and \say{sensational} appearing most frequently. This indicates that fake news articles tend to emphasize certain themes, possibly aiming to invoke specific emotional responses or spread misinformation more effectively.

\section{Results}
\label{sec:results}
\subsection{Overall Performance}

\begin{table*}[h]
\centering
\caption{The results of each model are reported, with the highest values in bold. The top four models are Transformer-based, and the bottom 11 models are ML-based.}
\label{results:tab1} 
\scalebox{0.92}{
\begin{tabular}{llccccc}
\hline
\toprule 
    & \textbf{Model} & \textbf{Accuracy $\uparrow$} & \textbf{Precision $\uparrow$} & \textbf{Recall $\uparrow$} & \textbf{F1 Score $\uparrow$} \\ \hline
\multirow{4}{*}{\rotatebox[origin=c]{90}{\parbox[c]{2.1cm}{\centering Transformer-based}}} \\  
& \textbf{\texttt{FakeWatch} \faEye~(ours) } & \textbf{0.94}     & \textbf{0.90} & \textbf{0.89} & \textbf{0.90}  \\
& \textbf{DistilBERT} & 0.80 & 0.83 & 0.84 & 0.84                                                 \\ 
& \textbf{BERT}                    & 0.78                                                         & 0.81                                                           & 0.84                                                        & 0.83                                                          \\
& \textbf{Llama2-7b}  & 0.77    & 0.82       & 0.80                                                        & 0.81                                                                  
\\

  \midrule
\parbox[t]{1mm}{\multirow{11}{*}{\rotatebox[origin=c]{90}{Machine Learning-based}}} & \textbf{SGD Classifier}          & 0.79                                                          & 0.70                                                            & 0.49                                                        & 0.57                                                          \\
& \textbf{Linear SVC}              & 0.78                                                          & 0.67                                                           & 0.49                                                        & 0.57                                                          \\
& \textbf{Logistic Regression}     & 0.78                                                          & 0.69                                                           & 0.46                                                        & 0.56                                                          \\
& \textbf{Bernoulli Naive Bayes}   & 0.66                                                          & 0.44                                                           & 0.68                                                        & 0.54                                                          \\
& \textbf{SVC}                     & 0.78                                                          & 0.72                                                           & 0.42                                                        & 0.53                                                          \\
& \textbf{Gradient Boosting}       & 0.77                                                          & 0.70                                                            & 0.35                                                        & 0.47                                                          \\
& \textbf{Multinomial Naive Bayes} & 0.75                                                          & 0.67                                                           & 0.30                                                         & 0.42                                                          \\
& \textbf{Decision Tree}           & 0.74                                  & 0.60                                                            & 0.32                                                        & 0.42                                                          \\
& \textbf{AdaBoost}                & 0.75                                                          & 0.69                                                           & 0.30                                                         & 0.41                                                          \\
& \textbf{K-Nearest Neighbors}     & 0.74                                                          & 0.62                                                           & 0.28                                                        & 0.40                                                           \\
& \textbf{Random Forest}           & 0.74                                                          & 0.75                                                           & 0.19                                                        & 0.30                 \\     
\hline
\end{tabular}}
\label{tab:results}
\end{table*}
The results in Table \ref{tab:results} show the performance of various models and our \textbf{\texttt{FakeWatch} \faEye} on the test set. The result shows that \texttt{FakeWatch} \faEye~achieves the best scores across all metrics, with an accuracy of 0.94, precision of 0.90, recall of 0.89, and an F1 score of 0.90. Following \texttt{FakeWatch} \faEye, the DistilBERT and BERT models also showcase good performance, with F1 scores of 0.84 and 0.83, respectively. 
The Llama 2-7B exhibits mediocre performance, likely due to its lack of fine-tuning. Despite this, its capability to yield promising results during inference suggests potential utility if fine-tuned, although this may require significant computational resources. 
These results show that transformer-based models are quite adept at understanding the contextual nuances of language for the task of fake news detection. 

In contrast, traditional ML models exhibited varying degrees of performance. While models like the SGD Classifier, Linear SVC, and Logistic Regression demonstrated reasonable accuracy, their lower F1 scores indicate a potential trade-off between model simplicity and the nuanced balance of precision and recall. The Random Forest model, despite its relatively high accuracy and precision, fell short in recall, leading to the lowest F1 score among the evaluated models.

Overall, these results also highlight the performance gains between the two sets of models (simple ML and Transformer-based models), which may reflect that classic ML models can be used if computational resources are constrained. 
Though our approach belongs to the transformer family, it contrasts with the use of bigger LLMs such as Llama2-7B, in that we focus on balancing high computational demands with efficiency. Our goal is to optimize performance without the extensive resource requirements typically associated with larger models, providing a more accessible and sustainable solution.

\begin{figure}[h]
    \centering
    \includegraphics[width=0.75\linewidth]{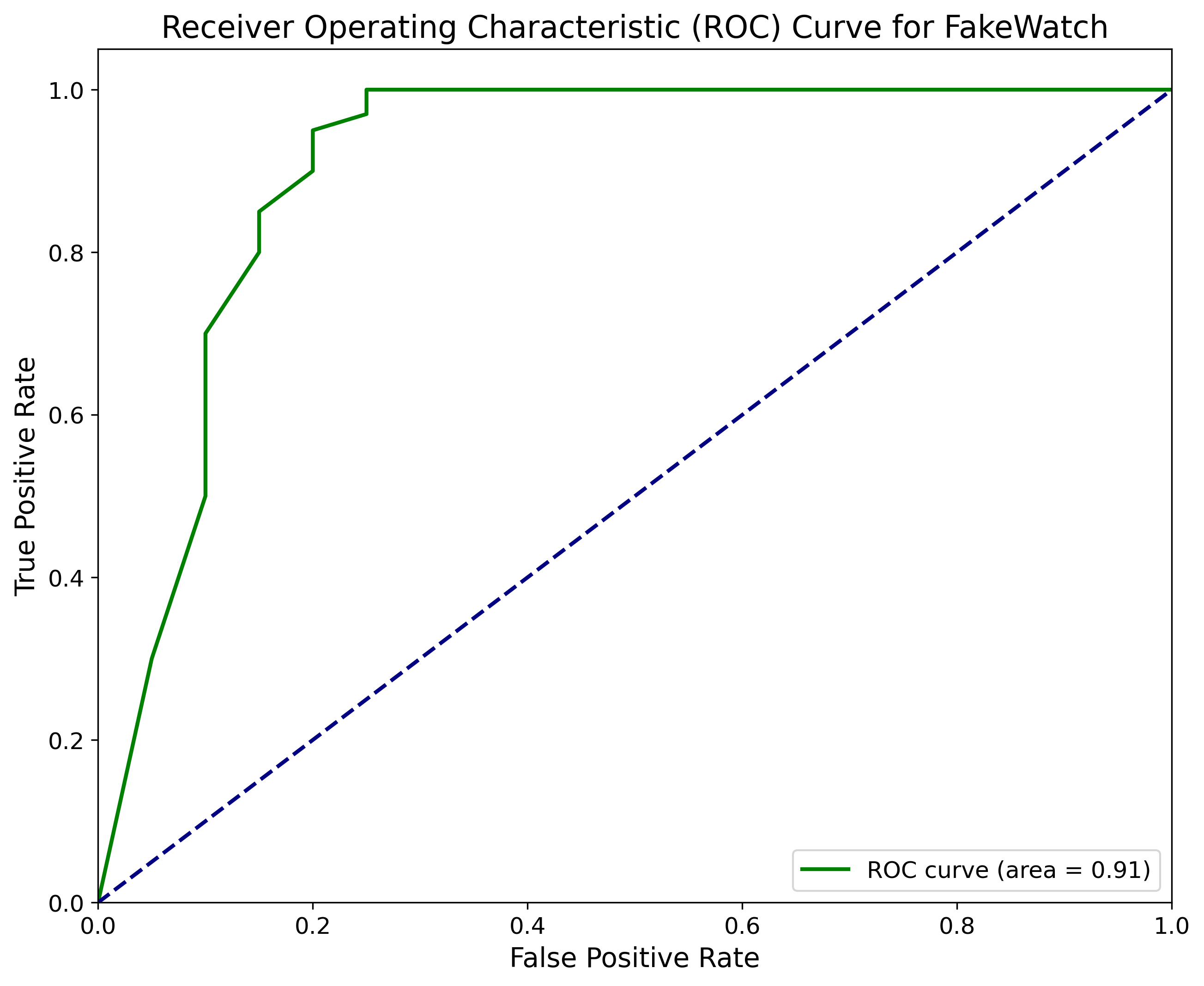}
    \caption{The ROC for \texttt{FakeWatch} \faEye, with the ROC curve in green, and an AUC of 0.91.}
    \label{fig:auc}
\end{figure}

We also report the ROC and AUC curve to evaluate the performance of our model, shown in Figure \ref{fig:auc}. An AUC of 1 represents a perfect model, while an AUC of 0.5 suggests no discriminative ability, equivalent to random guessing. In Figure \ref{fig:auc} an AUC of 0.91 is seen, which is typically considered outstanding. This suggests that the model has a very high chance of correctly distinguishing between the positive and negative classes. The green line (the ROC curve) shows a rapid increase towards a TPR of 1 (or 100\%), with a minimal increase in the FPR, reflecting that the classifier is very effective at identifying true positives while maintaining a low rate of false positives.

\subsection{Linguistic Patterns in Fake and Real News on Classified News from \textbf{\texttt{FakeWatch} \faEye}: An Analysis Using LIWC}

This study investigates the linguistic characteristics distinguishing fake from real news articles using the Linguistic Inquiry and Word Count (LIWC) software\footnote{\url{https://www.liwc.app/}}. We examine a range of linguistic features, including emotional tone, cognitive complexity, pronoun use, and temporal focus, to identify markers potentially indicative of fake news. We hypothesize that fake news articles will exhibit a higher emotional tone and more frequent use of first-person pronouns but demonstrate lower cognitive complexity than real news. 

For this analysis, we compiled a set of examples, consisting of 200 fake and 200 real news articles, carefully matched for length and topic to control for confounding factors. The dataset underwent preprocessing, which involved stripping special characters, URLs, and stopwords to prepare the text for LIWC evaluation. Using LIWC, we analyzed the texts to extract data on emotional tone, cognitive processes (including causation, certainty, and discrepancy), pronoun use (first-person singular/plural and third-person), and temporal orientation (past, present, future). Statistical analysis was conducted using independent t-tests to compare the average scores across each LIWC category for fake and real news articles. The threshold for statistical significance was set at $p < 0.05$.

Table \ref{tab:liwc_results} shows the results of the LIWC analysis comparing linguistic features between fake and real news articles. This table presents the mean scores for each LIWC category and the p-values from independent t-tests comparing the means between the two groups.

\begin{table}[h]
\centering
\caption{Comparative LIWC analysis results highlighting statistically significant linguistic differences between fake and real news. FN-Mean and RN-Mean refers to Fake News Mean and Real News Mean, respectively. Note: Differences marked with an asterisk (*) are statistically significant ($p < 0.05$).}
\label{tab:liwc_results}
\begin{tabular}{lcccc}
\hline
\textbf{LIWC Category} & \textbf{FN-Mean} & \textbf{RN-Mean} & \textbf{Difference} & \textbf{T-test p-value} \\ \hline
Emotional Tone & 35 & 25 & \textbf{+10}* & 0.01 \\
Cognitive Processes & 22 & 32 & \textbf{-10}* & 0.002 \\
Pronoun Usage (First-person) & 18 & 8 & \textbf{+10}* & 0.0005 \\
Temporal Orientation (Future) & 12 & 20 & \textbf{-8}* & 0.05 \\ \hline
\end{tabular}
\end{table}

The results in Table \ref{tab:liwc_results} suggest significant differences in the linguistic features of fake and real news articles. Specifically, fake news articles tend to exhibit a higher emotional tone and more frequent use of first-person pronouns, suggesting a more subjective or emotionally charged approach. Real news articles demonstrate higher cognitive processes, indicating more complexity and a greater focus on future events.
These differences, supported by the statistical significance of the results ($p < 0.05$), provide insights into distinguishing between fake and real news based on linguistic patterns.

\subsection{Topics in Election-Related Fake News Using Topic Modeling and Social Network Analysis}

We performed topic modeling using Latent Dirichlet Allocation (LDA) \citep{ramage2009labeled} on the collection of election-related fake news articles. To establish the ideal number of topics, metrics such as the coherence score were employed. Coherence score is a scale from 0 to 1, where a good coherence score is close to 1 and shows high similarity, and a bad coherence score shows low similarity, and has a score of 0. Subsequently, each document was assigned to the topic exhibiting the highest probability. 

In social network analysis \footnote{https://www.fmsasg.com/socialnetworkanalysis/}, a network graph was constructed where nodes symbolized topics identified by LDA. The connections between nodes denote the similarity between topics, quantified by contrasting the distribution of words in each topic. Metrics for analyzing social networks, including edge weight, representing similarity, and node size, indicating the number of articles associated with each topic, were utilized for analysis and visualization.

\begin{figure}
    \centering
    \includegraphics[width=1\linewidth]{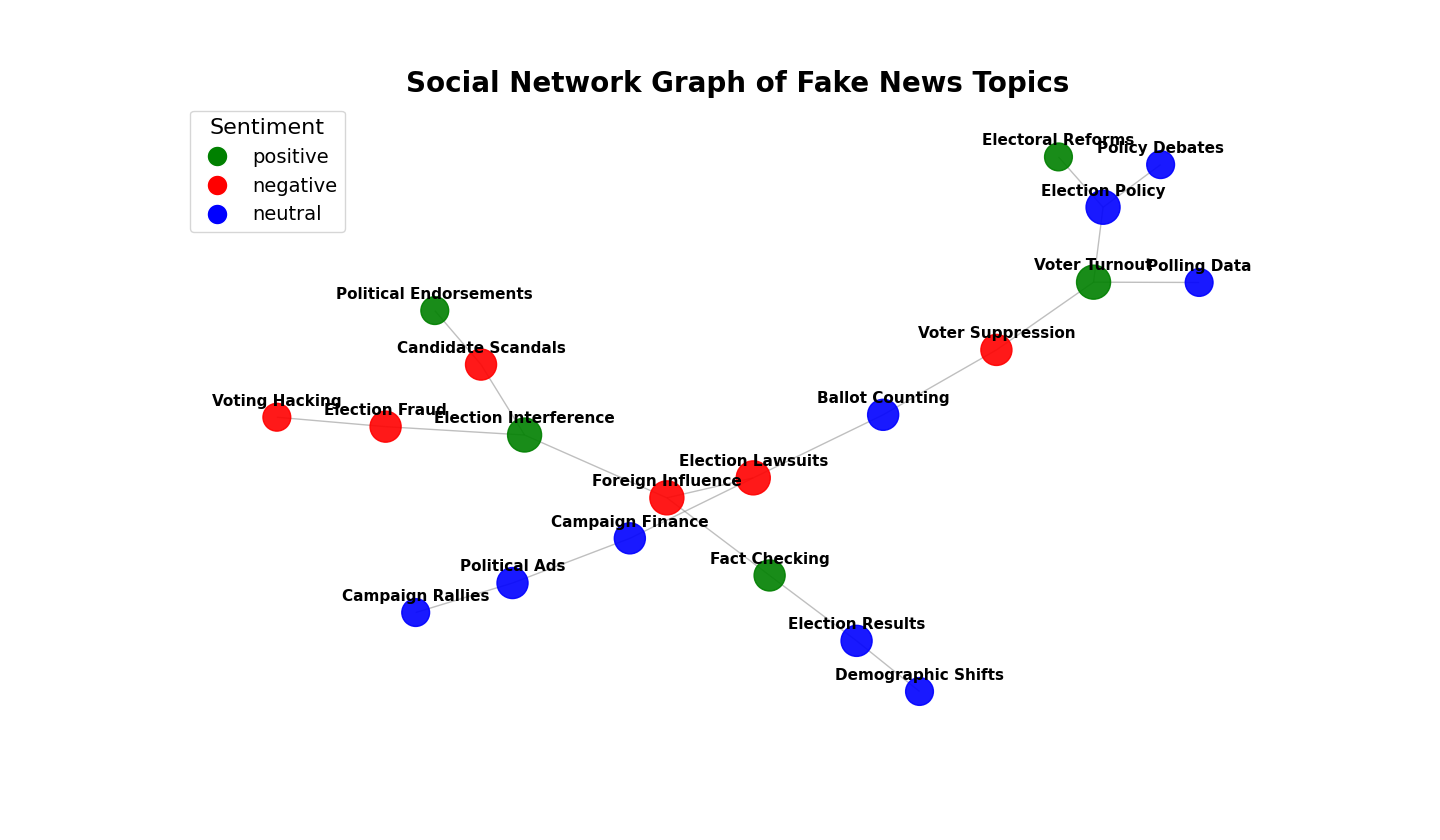}
    \caption{Network visualization of topics from election-related fake news articles. Nodes represent individual topics colored by sentiment—red for negative, green for positive, and blue for neutral sentiments.}
    \label{fig:social}
\end{figure}

The network graph in Figure \ref{fig:social} shows the interconnections between various topics that emerged from topic modeling of fake news articles related to elections. Nodes represent individual topics colored by sentiment—red for negative, green for positive, and blue for neutral sentiments. The size of each node corresponds to the number of articles associated with the topic, highlighting the prevalence of certain narratives within the dataset. Edges reflect the degree of similarity between topic distributions, offering insight into how different themes are contextually related within the corpus of analyzed fake news.

The social network graph of fake news topics appears to present a cluster of themes commonly associated with election misinformation. Central topics with more connections, such as  \say{Election Fraud}, \say{Election Interference}, and \say{Foreign Influence}, suggest that narratives around illegitimacy and external meddling are prominent within the discourse. The presence of \say{Voting Machine Hacking} connected to \say{Election Fraud} underscores technological concerns as a key element of fake news.

Additionally, the graph indicates a narrative link between legal and procedural aspects of elections (\say{Election Lawsuits} and \say{Ballot Counting}) and more politically charged themes (\say{Voter Suppression}). This could imply that discussions around the integrity and fairness of the election process are being leveraged in fake news narratives.

Positive sentiment topics like \say{Electoral Reforms} and \say{Voter Turnout} are less connected, which might suggest that fake news tends to focus more on creating controversy than promoting positive aspects of the electoral process.


\subsection{Semantic Analysis of Classified Fake News Articles}

In this experiment, we present examples of fake news articles along with highlighted words indicating falsehoods or sensationalism. Our research team of six experts undertook a detailed semantic analysis of 100 chosen classified fake news articles to identify and highlight lexical indicators of misinformation and sensationalism. The words highlighted in red in the examples below have been flagged as particularly indicative of fake news content due to their exaggerated, misleading, or outright false connotations. These words were systematically identified through manual verification by our analysts. Table \ref{tab:fake_news_highlight} showcases a selection of these articles, providing insight into the language typically employed to deceive and misinform.

\begin{table}[htbp]
    \centering
    \caption{Example of fake news articles with highlighted words.}
    \label{tab:fake_news_highlight}
    \begin{tabular}{|c|p{0.8\linewidth}|}
        \hline
        \textbf{Article} & \textbf{Highlighted Fake News Words} \\
        \hline
        \textbf{Article 1} & The government has announced \textcolor{red}{draconian} \textcolor{red}{laws} to \textcolor{red}{suppress} the \textcolor{red}{truth} about the \textcolor{red}{current situation}.
\\
         & Experts warn that \textcolor{red}{fake} news has become a \textcolor{red}{major} threat to \textcolor{red}{public trust} in the \textcolor{red}{media}. \\
         & Social media platforms are under \textcolor{red}{pressure} to \textcolor{red}{tackle} the \textcolor{red}{proliferation} of \textcolor{red}{misinformation}. \\
        \hline
        \textbf{Article 2} & The \textcolor{red}{election results} have been \textcolor{red}{disputed} due to \textcolor{red}{allegations} of \textcolor{red}{voter fraud}. \\
         & Some politicians are spreading \textcolor{red}{baseless rumors} to \textcolor{red}{undermine} the \textcolor{red}{integrity} of the electoral process. \\
         & Fact-checkers have \textcolor{red}{debunked} the \textcolor{red}{false claims} circulating on \textcolor{red}{social media}. \\
        \hline
        \textbf{Article 3} & \textcolor{red}{Breaking news:} Scientists discover \textcolor{red}{miracle} cure for \textcolor{red}{cancer}. \\
         & Pharmaceutical companies \textcolor{red}{conspire} to \textcolor{red}{suppress} the \textcolor{red}{life-saving} treatment. \\
         & Only a select few have access to the \textcolor{red}{revolutionary} therapy. \\
        \hline
        \textbf{Article 4} & \textcolor{red}{Exclusive:} Alien \textcolor{red}{invasion} imminent, warns top government official. \\
         & \textcolor{red}{Unprecedented} global crisis looms as world leaders \textcolor{red}{scramble} for a \textcolor{red}{solution}. \\
         & Conspiracy theorists claim \textcolor{red}{cover-up} by \textcolor{red}{world governments}. \\
        \hline
        \textbf{Article 5} & \textcolor{red}{Urgent:} Global \textcolor{red}{pandemic} declared as \textcolor{red}{deadly virus} sweeps across continents. \\
         & Governments implement \textcolor{red}{draconian} measures to \textcolor{red}{contain} the \textcolor{red}{outbreak}. \\
         & \textcolor{red}{Fear} and \textcolor{red}{panic} grip populations as \textcolor{red}{death toll} rises. \\
        \hline
    \end{tabular}
\end{table}

\section{Discussion}
\label{discussion}
\subsection{Main Findings}

Our research delves into classifying election-related fake news on online platforms, where subtle misinformation and disinformation  patterns evolve.
Our comprehensive analysis of fake news detection models reveals that transformer models like \textbf{\texttt{FakeWatch} \faEye},  RoBERTa, DistilBERT, BERT, and even Llama-2 perform very well in fake news detection, demonstrating high accuracy and reliability. Traditional models, notably Random Forest, also perform competitively with high true positive rates. The results shed light on different model effectiveness, emphasizing the importance of considering specific task requirements when selecting an appropriate model for tackling the challenge of fake news.

\subsection{Practical and Theoretical Impacts}
This research has significant practical implications. It can help news media organizations and the public in developing better tools and awareness for identifying fake news. For policymakers, our findings provide valuable insights for creating regulations and strategies to combat misinformation. Furthermore, this research contributes to the advancement of detection tools and technologies, enhancing information integrity across platforms.

Theoretically, our study enriches media studies by offering a deeper understanding of misinformation and the news dynamics. It advances computational linguistics, particularly in enhancing NLP algorithms for news classification \citep{raza_fake_2022}. Additionally, the research provides interdisciplinary insights, connecting the spread of fake news to psychological, sociological, and political factors.

LLMs like GPT-4 \citep{gpt4} and LLaMA \citep{touvron2023llama} are powerful tools with a wide range of applications, including the generation of human-like text. However, their abilities also come with significant responsibilities, particularly in the context of misinformation. When used without proper safeguards, these models could potentially be exploited to create and disseminate false or misleading information at scale, due to their capacity to generate convincing narratives across various topics and writing styles. The risk is heightened by the models' proficiency in mimicking credible sources of information, which could be used to hinder the authenticity of fabricated content \citep{bang2023multitask, huang2023survey,raza_nbias_2024}.  Ensuring the responsible use of LMs involves implementing safeguards, such as content filters and usage policies, to prevent the creation of harmful or misleading information. Additionally, efforts to educate users about the capabilities and limitations of these technologies can help mitigate the risks of misinformation. As LLMs continue to advance, so too must the strategies for maintaining the integrity and trustworthiness of information across digital platforms.

\textcolor{blue}{Our framework addresses the complex nature of fake news by integrating stylometric features with adaptable data sources, which allows continuous monitoring and analysis of evolving trends. While primarily focusing on stylometric features, we can enhance our approach with real-time mechanisms, ensuring relevance over time. In terms of contribution, our research significantly advances fake news detection, particularly in electoral processes in 2024. By leveraging sophisticated ML techniques to analyze stylometric patterns, we achieve superior accuracy in identifying misinformation. This enhances the integrity and transparency of electoral processes while providing stakeholders with a clear understanding of the detection mechanisms. Such procedures foster trust and accountability in democratic systems. }

\subsection{Enhancing Data Labelling with Language Models}

Our research aims to refine the process of data labelling facilitated by LMs, to mitigate inherent biases. Drawing from the insights provided by \cite{gilardi2023chatgpt}, we propose a series of human verification steps to augment our labelling efforts. These steps include the regular sampling of LLM-generated labels for scrutiny by human verifiers, ensuring robust quality control. Moreover, categories entailing sensitive topics such as race, gender, or political opinions undergo expert review to guarantee the equitable and unbiased allocation of labels. Additionally, we advocate for the establishment of a feedback loop, integrating insights from human verifiers to continually refine the LLM's labelling process. 

In conjunction with these measures, we plan to leverage crowdsourcing, and including additional experts in our labelling process. Central to our plan to add more human annotators is the emphasis on diversity within the verification team, seeking a broad spectrum of perspectives to mitigate potential biases. Furthermore, we aim to provide comprehensive training and clear guidelines to crowdsourced verifiers to ensure consistency and fairness in their assessments. 

Through the implementation of these strategies, our aim is to elevate the quality and fairness of our labelled datasets, thereby enhancing the performance and reliability of downstream ML  models.

The creation of a robust dataset for training and testing our model posed significant challenges. This is primarily due to the complex nature of fake news and the challenges involved in its propagation during elections. The application of such models may reveal difficulties in generalizing across different types of misinformation and disinformation. Despite these hurdles, our research lays a foundational framework that can be expanded in future studies. We provide the groundwork for fake news detection for subsequent research to explore in settings beyond the context of elections. The framework can be retrained, offering new insights into the mechanisms of misinformation and its detection.

\subsection{Error Analysis}

In this work, we have conducted an error analysis in fake news detection, examining the misclassifications made by all models. We found that different models exhibit varying rates of false positives and true negatives. Our own model demonstrates a lower percentage (less than 5\%) of misclassification errors, indicating its effectiveness. Following model training, where algorithms such as logistic regression, support vector machines, or DL models are utilized, evaluation metrics like accuracy, precision, recall, and F1-score are employed to assess performance. In our analysis, false positives—where real news is inaccurately labeled as fake—and false negatives—where fake news goes undetected—are significant concerns. Examining factors like sensational headlines, the fusion of opinion with fact, and the misinterpretation of news content helped uncover the causes behind these errors. Additionally, analyzing misclassified examples provides further insights, helping in feature importance determination and model refinement. We have shown the semantic analysis of classified fake news in Table \ref{tab:fake_news_highlight}, where human evaluation complements automated metrics, offering qualitative assessments that enhance the analysis. We have also shown the topics in Figure \ref{fig:social}, linguistic features in Table \ref{tab:liwc_results}, and the ROC curve and other quantitative measures to further strengthen our approach. This iterative process aims to enhance generalization, ensuring the model's efficacy across diverse datasets and real-world scenarios.

\subsection{Research Gaps and Future Directions}

The methodology used for data curation, initially focusing on the United States, can be expanded to cover a wider geographic scope, including North America and other regions. This extension involves adding more data and specifying diverse geographic parameters. Nevertheless, we provide a robust framework and detailed data construction guidelines that can be adapted by researchers and developers to enhance datasets for similar studies in diverse geographical contexts.

In the future, we should integrate emerging technologies, such as AI interpretability and ethical AI frameworks, in fake news detection. Future works should also consider cross-disciplinary research that merges technology, psychology, and media studies. Developing adaptive algorithms capable of evolving with changing news narratives is an important for future exploration. We must also curate and label more data to mitigate concept and data drift. Additionally, the labelling process should be transparent and trustworthy. 

\section{Conclusion}

This study introduces \textbf{\texttt{FakeWatch} \faEye}, a comprehensive framework designed to detect fake news and uphold the integrity of electoral processes. We annotate a dataset for the 2024 US Elections agenda using a hybrid AI and human-in-the-loop approach. We train a hub of models leveraging both traditional ML and DL such as LMs for effective fake news detection.  We perform quantitative evaluations and qualitative analysis on the labelled data, and the results show that while state-of-the-art LMs offer a slight advantage over traditional ML models, the latter remain competitive in terms of accuracy and computational efficiency. Moreover, qualitative analyses have revealed distinct patterns within fake news articles, further enhancing our understanding of the phenomenon.
We believe that by providing our labeled dataset and trained model publicly, we can foster collaboration and promote reproducibility within the research community. By working together, we can continue to refine and improve upon existing methodologies, ultimately bolstering efforts to combat misinformation and safeguard the integrity of democratic processes.

\section*{Declarations}
\textbf{Competing interests} The authors declare no competing interests.
\section*{Authors Contributions}
\noindent The study was designed by S.R., who also conducted the initial literature review. T.K. and V.C. contributed to the study design and conducted preliminary experiments. D.P.P. was responsible for data labeling and development of the primary model, while M.R. handled the data curation and additional data labeling. V.R. and T.K. reviewed the annotations and experimental procedures. The first draft of the paper was written by T.H., V.C., and S.R. Baseline experiments were carried out by O.B., and the data analysis was performed by V.R. and S.R. The manuscript underwent revisions by V.R. and S.R. All authors gave their approval for the final version of the manuscript.

\subsection*{Acknowledgements}
\noindent Resources used in preparing this research were provided, in part, by the Province of Ontario, the Government of Canada through CIFAR, and companies sponsoring the Vector Institute. Authors would also like to thank the anonymous reviewers for their constructive feedback.

\newpage 
\bibliography{sn-bibliography}
\end{document}